\documentclass[sigconf,screen]{acmart}

\usepackage{microtype}
\usepackage{caption}
\usepackage{graphicx}
\usepackage{booktabs} % for professional tables
\usepackage{graphicx}
\usepackage{amsmath}
\usepackage{hyperref}
\usepackage{multirow}
\usepackage{subcaption}
\usepackage{textcomp}
\usepackage{booktabs}
\usepackage{colortbl}  %彩色表格需要加载的宏包
\usepackage{xcolor}
\usepackage{array}  
\usepackage{booktabs}
\usepackage[table]{xcolor}
\usepackage{colortbl}

\definecolor{best}{RGB}{231,209,255}     % 深紫（best）
\definecolor{second}{RGB}{243,229,255}   % 中紫
\definecolor{third}{RGB}{250,242,255}    % 

\usepackage{pifont}
\newcommand{\cmark}{\textcolor{green!60!black}{\ding{51}}} % ✓
\newcommand{\xmark}{\textcolor{red!70!black}{\ding{55}}}   % ✗ 

\AtBeginDocument{%
  }

\setcopyright{acmlicensed}
\copyrightyear{2018}
\acmYear{2018}
\acmDOI{XXXXXXX.XXXXXXX}
\acmConference[Conference acronym 'XX]{Make sure to enter the correct
  conference title from your rights confirmation email}{June 03--05,
  2018}{Woodstock, NY}
\acmISBN{978-1-4503-XXXX-X/2018/06}

\begin{document}

\title{Learning to Synergize Semantic and Geometric Priors for Limited-Data Wheat Disease Segmentation}

\author{
Shijie Wang$^1$, Zijian Wang$^1$, Yadan Luo$^1$, Scott Chapman$^1$, Xin Yu$^2$, Zi Huang$^1$ \\
   \textsuperscript{1} The University of Queensland, Australia  
   \textsuperscript{2} The University of Adelaide, Australia   }
\renewcommand{\shortauthors}{Wang et al.}

\begin{abstract}

Wheat disease segmentation is fundamental to precision agriculture but faces severe challenges from significant intra-class temporal variations across growth stages. Such substantial appearance shifts make collecting a representative dataset for training from scratch both labor-intensive and impractical. To address this, we propose SGPer, a \underline{S}emantic-\underline{G}eometric \underline{P}rior Syn\underline{er}gization framework that treats wheat disease segmentation under limited data as a coupled task of disease-specific semantic perception and disease boundary localization. Our core insight is that pretrained DINOv2 provides robust category-aware semantic priors to handle appearance shifts, which can be converted into coarse spatial prompts to guide SAM for the precise localization of disease boundaries. Specifically, SGPer designs disease-sensitive adapters with multiple disease-friendly filters and inserts them into both DINOv2 and SAM to align their pretrained representations with disease-specific characteristics. To operationalize this synergy, SGPer transforms DINOv2-derived features into dense, category-specific point prompts to ensure comprehensive spatial coverage of all disease regions. To subsequently eliminate prompt redundancy and ensure highly accurate mask generation, it dynamically filters these dense candidates by cross-referencing SAM’s iterative mask confidence with the category-specific semantic consistency derived from DINOv2. Ultimately, SGPer distills a highly informative set of prompts to activate SAM’s geometric priors, achieving precise and robust segmentation that remains strictly invariant to temporal appearance changes. Extensive evaluations demonstrate that SGPer consistently achieves state-of-the-art performance on wheat disease and organ segmentation benchmarks, especially in data-constrained scenarios.

\end{abstract}



\maketitle

\section{Introduction}
Wheat is one of the most important staple crops worldwide, providing a major source of dietary calories for human populations \cite{DBLP:journals/cea/FanLYLWZZCMGX26, DBLP:journals/cea/JiangFZYFCTZCL26}. However, its yield and quality are continually threatened by diverse diseases, which cause substantial economic losses and pose serious risks to global food security. In this context, accurate disease segmentation \cite{DBLP:journals/cea/LeiYZYLHQH26, DBLP:conf/adc/WangWLZY25, DBLP:journals/cea/ZhangHLWLLM25} is of great practical significance, as it enables precise disease localization and supports timely disease monitoring, diagnosis, and intervention in modern wheat production systems.

Despite its importance, wheat disease segmentation in real-world field environments remains highly challenging \cite{DBLP:conf/adc/WangWLZY25, DBLP:conf/adc/WeiYCCH25, DBLP:conf/iccvw/Fortin23}. A major difficulty is that the visual appearance of the same disease can change remarkably across different infection stages. Meanwhile, different disease categories often exhibit highly similar textures, colors, and lesion patterns, which further increases the demand for disease-specific representation learning. 
%As a result, constructing a representative dataset requires collecting images that capture diverse disease appearances across all infection stages, which is labor-intensive and challenging in practice. This limitation leads to insufficient coverage of diverse disease variations and forces models to rely on scarce annotated data.
%Consequently, existing datasets fail to capture the full spectrum of temporal appearance variations. This limited coverage forces models to rely on scarce annotated data, significantly restricting their ability to learn generalizable representations.
Consequently, the extreme diversity of temporal appearances is underrepresented in existing datasets, as collecting images covering all disease stages is labor-intensive and challenging in practice. This limitation results in insufficient coverage of disease variations and forces models to rely on scarce annotated data. 
Therefore, effective wheat disease segmentation requires robustness to large intra-class temporal variations while accurately distinguishing visually similar categories under limited supervision.

Recent advances in foundation models \cite{DBLP:conf/cvpr/SunCZZCZDWL24, DBLP:conf/iccvw/ChenZDCWZLSZM23, DBLP:journals/corr/abs-2304-07583} offer new opportunities for accurate wheat disease segmentation under limited supervision. 
In particular, the Segment Anything Model (SAM) demonstrates strong zero-shot transferability and effective task adaptation across diverse domains, offering a promising pathway to alleviate the reliance on large-scale annotated data. However, while SAM provides powerful geometric priors for accurate boundary localization, its lack of disease-aware semantic perception limits its zero-shot performance on wheat diseases with complex appearances. 
Moreover, although interactive prompting for SAM can partially mitigate this issue, manual guidance remains impractical: sparse prompts provide insufficient geometric cues to fully activate SAM’s capabilities, resulting in inaccurate segmentation of irregular lesions, whereas dense prompting is prohibitively labor-intensive.
This raises a critical question: can SAM’s geometric priors be automatically guided by a model with richer semantics, rather than relying on dense prompts provided by users? DINOv2~\cite{oquab2024dinov2}, benefiting from large-scale self-supervised pre-training, naturally bridges this gap, allowing it to readily capture highly varying disease patterns. 
Therefore, how to effectively translate DINOv2’s robust semantic representations into dense, high-quality point prompts to automatically unlock SAM’s geometric potential without human intervention is worthy of investigation.
%Therefore, a promising solution is to automatically transform DINOv2’s semantic priors into dense prompts, thereby activating SAM’s geometric decoding for precise and labor-efficient disease segmentation.

To this end, we propose SGPer, a Semantic–Geometric Prior Synergization framework that establishes a synergistic paradigm between between DINOv2 and SAM. It leverages DINOv2 to provide robust, category-aware semantic priors that are resilient to appearance shifts across growth stages, while using these semantic priors to guide SAM in exploiting its geometric priors for precise localization of irregular lesion boundaries.
Our core insight is to formulate wheat disease segmentation under limited data as a tightly coupled task of disease-specific semantic perception and fine-grained boundary localization.
Through this tight coupling of semantic and geometric priors deived from foundation models, SGPer achieves accurate segmentation of complex wheat diseases while drastically reducing the reliance on extensive training annotations.
%Motivated by the above analysis, we propose SGPer, a Semantic-Geometric Prior Synergization framework that formulates wheat disease segmentation under limited data as a tightly coupled task of disease-specific semantic perception and disease boundary localization. Our core insight is to establish a synergistic paradigm between foundation models: leveraging DINOv2 to provide robust, category-aware semantic priors that resist appearance shifts across growth stages, while guiding SAM to utilize its geometric priors for the precise localization of irregular lesion boundaries. By seamlessly integrating these complementary pretrained capabilities, SGPer achieves accurate segmentation of complex wheat diseases while drastically reducing the reliance on extensive training annotations.

Technically, SGPer achieves this synergy through three key operations. First, to bridge the domain gap between generic pretrained representations and wheat disease imagery, we design disease-sensitive adapters equipped with multiple disease-friendly filters. By inserting them into both DINOv2 and SAM, we effectively align their general-purpose pretrained representations with the specific characteristics of wheat diseases. Second, to operationalize the semantic-to-geometric guidance, SGPer transforms the refined DINOv2 features into dense, category-specific point prompts, ensuring exhaustive spatial coverage of all disease regions in the images. Finally, to remove low-quality points and retain a minimal yet highly effective set of prompts, we introduce a synergistic pruning mechanism that dynamically filters dense candidates by cross-referencing SAM’s iterative mask confidence with category semantics derived from DINOv2. Through this cohesive pipeline, SGPer effectively distills a highly informative set of prompts that activates SAM’s geometric priors, enabling precise and robust segmentation that remains invariant to temporal appearance changes.

Our main contributions are summarized below: 
\begin{itemize}
    \item %To the best of our knowledge, this is one of the first attempts to address wheat disease segmentation under substantial cross-stage appearance variation and limited supervision in field environments.
    To the best of our knowledge, we are the first attempt to formulate wheat disease segmentation under limited data as a tightly coupled task of disease-specific semantic perception and disease boundary localization.

    \item We propose SGPer, a novel framework that leverages DINOv2 to provide robust, category-aware semantic priors resilient to appearance variations across growth stages, while guiding SAM to exploit its geometric priors for precise localization of disease boundaries.

    \item Extensive experiments demonstrate that SGPer achieves state-of-the-art performance on both wheat disease and organ segmentation benchmarks, particularly excelling under limited training data.
\end{itemize}

\begin{figure*}
    \centering
    \includegraphics[width=1\linewidth]{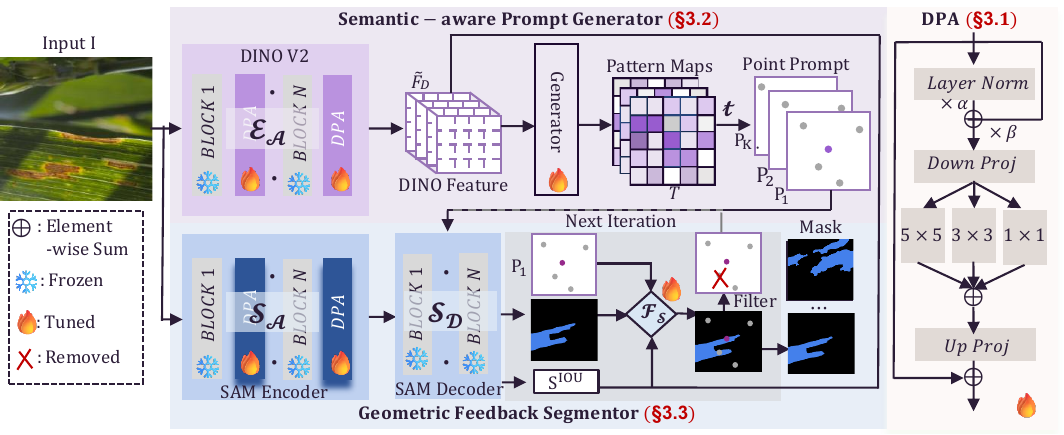}
    \caption{Framework illustration of Semantic-Geometric Prior Synergization. See \textcolor{red}{\S \ref{sec:method}} for more details.}
    \label{fig:method}
\end{figure*}

\section{Related Work}
\noindent\textbf{Plant disease segmentation.}
Wheat disease segmentation has gained increasing attention in agricultural research, as it plays a crucial role in containing disease spread and reducing yield losses.
Prior work~\cite{zenkl2022outdoor} indicates that outdoor wheat disease segmentation under uncontrolled conditions remains an open challenge, posing a significant barrier to fully automated agricultural systems. Existing studies also explore robotic platforms for disease detection to support downstream analysis and targeted intervention~\cite{singh2017detection}. Moreover, segmentation serves as a fundamental component in plant phenotyping, with applications including leaf counting~\cite{DBLP:conf/iccvw/AichS17}, ear counting~\cite{david2020global}, and multi-scale plant–soil segmentation, all of which facilitate plant growth monitoring and tracking.
However, existing works largely overlook the severe intra-class temporal variations that occur across different growth stages. This oversight leads to insufficient data diversity, severely limiting model generalization and degrading segmentation accuracy on rarely observed disease patterns. 
Therefore, we are the first to leverage semantic priors derived from pretrained DINOv2 and the geometric priors of SAM, reformulating wheat disease segmentation under limited data as a coupled task of disease-specific semantic perception and disease boundary localization.

\noindent\textbf{Segment Anything Model and Follow-Ups.}
Segment Anything Model (SAM)~\cite{kirillov2023segment} has emerged as a powerful category-agnostic foundation model, driven by interactive spatial prompts such as points, bounding boxes, and coarse masks. Its robust generalization capabilities have catalyzed advancements across diverse downstream tasks, including image matting \cite{DBLP:conf/cvpr/0003JS22}, video tracking \cite{DBLP:journals/corr/abs-2304-11968}, and medical image segmentation \cite{DBLP:conf/aaai/YueZH0L024}. However, when applied to highly specialized fields, SAM often struggles due to significant domain gaps. To mitigate this, recent approaches~\cite{DBLP:journals/corr/abs-2304-12620} integrate adapter tuning into SAM’s ViT encoder to tailor its representations for domain-specific visual patterns. Alternatively, methods like PerSAM~\cite{DBLP:conf/iclr/Zhang0GYP000024} and Matcher~\cite{DBLP:conf/iclr/LiuZL0WS24} propose training-free, one-shot adaptation frameworks to align SAM with new domains. Furthermore, HQ-SAM~\cite{DBLP:conf/nips/KeYDLTT023} introduces a learnable high-quality output token to enhance disease mask details while preserving the model's zero-shot generalizability.
Despite these extensive domain adaptations, the potential of foundation models for wheat disease segmentation remains largely unexplored. To bridge this gap, we present the first attempt to seamlessly couple the strengths of two foundation models for this task. Specifically, we leverage DINOv2 to provide robust, category-aware semantic priors, which are transformed into automated dense point prompts. This semantic guidance effectively directs SAM, activating its geometric priors for the precise localization of highly irregular lesion boundaries.

\noindent\textbf{Visual Representation Learning.}
Self-supervised vision transformers have shown strong transferability for dense visual understanding. DINO~\cite{caron2021emerging} reveals that self-supervised ViTs naturally encode object semantics and scene layout, while DINOv2~\cite{oquab2024dinov2} further learns robust image- and pixel-level visual features at scale. Related work has exploited large pretrained priors for dense prediction, for example by injecting pretrained semantic knowledge into segmentation pipelines~\cite{rao2022denseclip}. However, most existing methods treat these pretrained representations as generic backbones or auxiliary cues, rather than explicitly converting semantic priors into prompts that activate a promptable segmentation model. In contrast, our method uses DINOv2 to derive disease-aware point prompts for SAM and projects the refined masks back into the semantic embedding space, forming a closed semantic-geometric loop for limited-data wheat disease segmentation. 

\section{Method}
\label{sec:method}
As depicted in Fig.~\ref{fig:method}, the core of SGPer lies in the synergistic collaboration between pretrained DINOv2 and SAM. First, SGPer introduces lesion-sensitive adapters into both foundation models, aligning their general-purpose representations with the specific characteristics of wheat diseases. Next, the adapted DINOv2 transforms its enriched semantic features into dense, category-specific point prompts, ensuring comprehensive spatial coverage of the disease regions. Finally, SAM selectively processes these point prompts to activate its robust geometric priors, enabling the precise localization and disease boundary localization of complex wheat lesions.

\subsection{Disease Perception Adapter}
\label{sec:DPA}
While foundation models like DINOv2 and SAM offer robust pretrained priors, their generic feature representations inevitably fall short when confronted with the subtle and complex appearance of wheat diseases. To bridge this critical gap, we introduce a lightweight Disease Perception Adapter (DPA) into both architectures. Rather than catastrophically forgetting their rich pretrained knowledge, DPA effectively recalibrates the pretrained features toward disease-specific visual distributions. Ultimately, DPA serves as a key component that enhances DINOv2’s sensitivity to disease semantics for accurate dense prompt generation, while guiding SAM to better capture the intricate geometries of diseases for precise mask localization.

Given an input token feature $F$, we first calibrate it from a distributional perspective to align pretrained features from foundation models (e.g., DINOv2 and SAM) with wheat disease imagery.
Mathematically, the calibration step can be formulated as:
\begin{equation}
    F_E = \alpha \cdot \mathrm{LN}(F) + \beta \cdot F,
\end{equation}
where $\mathrm{LN}(\cdot)$ denotes Layer Normalization, and $\alpha$ and $\beta$ are learnable scaling factors. 

In addition, to explicitly capture disease representations across substantial appearance variations, we introduce a lightweight disease-aware bottleneck module implemented as an encoder–decoder structure. The encoder compresses the calibrated feature $F_E$ into a lower-dimensional latent space and applies parallel depth-wise convolutions for multi-granularity disease extraction. Subsequently, the decoder utilizes point-wise convolution to fuse these representations before restoring the original channel dimension. The overall transformation is concisely formulated as:
\begin{equation}
    \begin{split}
        F_M & = \mathrm{Down}(F_E) + \frac{1}{3}\sum_{i\in \{1,3,5\}}W_{d}^i \hat{\otimes} \mathrm{Down}(F_E), \\
        \hat{F} & = F + \mathrm{Up}\big(\sigma(F_M + W_{p} \overline{\otimes} F_M)\big),
    \end{split}
\end{equation}
where $\mathrm{Down}(\cdot)$ and $\mathrm{Up}(\cdot)$ denote the channel projection layers, $\sigma(\cdot)$ represents the GeLU activation, and $\hat{\otimes}$ along with $\overline{\otimes}$ indicate the depth-wise and point-wise convolutions parameterized by $W_{d}^i$ (with kernel sizes $i \in \{1,3,5\}$) and $W_{p}$, respectively. Ultimately, this adapter effectively mitigates the domain gap by utilizing a bottleneck branch equipped with feature recalibration and multiple disease-friendly filters to robustly perceive subtle disease semantics, while simultaneously relying on a residual identity connection to preserve rich pretrained knowledge

We insert this unified, lightweight DPA architecture into both DINOv2 and SAM to adapt their model-specific priors for wheat disease segmentation. Despite sharing the same structure, DPA plays distinct roles in the two branches. In DINOv2, it strengthens disease-aware semantics within dense visual features, making disease regions more distinguishable and providing reliable cues for dense prompt generation. In SAM, it steers the prompt-driven segmentation process toward disease-relevant geometric structures, allowing it to better focus on disease extents and boundary details across varying appearances. Consequently, this unified adapter design exhibits strong generality, serving complementary purposes in both semantic enhancement and geometric guidance. %This unified adaptation further facilitates the collaboration between DINOv2 and SAM in SGPer, where disease-aware semantic cues support prompt generation and SAM-refined masks provide more reliable spatial evidence for subsequent semantic classification.

\subsection{Semantic-aware Prompt Generator}
\label{sec:ppg}
A major challenge in wheat disease segmentation lies in extreme intra-class variation across infection stages and high inter-class similarity in local textures. Consequently, relying on sparse prompts for SAM often fails to accurately segment diseases with diverse appearances, while manual dense prompting is impractical. To address this limitation, we introduce a Semantic-aware Prompt Generator (SPG), which serves as a key bridge that explicitly translates the rich disease semantics extracted by DINOv2 into spatially accurate dense point prompts tailored for SAM’s decoding process.

Given an input image $I \in \mathbb{R}^{H \times W \times 3}$, we extract its dense visual features using a frozen DINOv2 encoder coupled with our Disease Perception Adapter (DPA). Representing this augmented encoder as $\mathcal{E}_\mathcal{A}(\cdot)$, the adapted feature $\tilde{F}_{D} \in \mathbb{R}^{\frac{H}{s} \times \frac{W}{s} \times C}$ is obtained via:
\begin{equation}
    \tilde{F}_{D} = \mathcal{E}_\mathcal{A}(I),
\end{equation}
where $s$ is the patch stride and $C$ indicates the channel dimension. By explicitly integrating disease-specific details, $\tilde{F}_{D}$ yields a highly calibrated representation. This ensures robust feature discrimination against complex appearance variations, allowing SGPer to effectively identify wheat diseases.

To generate point prompts, PPG transforms the calibrated features to identify disease-relevant pattern maps $T\in \mathbb{R}^{K \times \frac{H}{s} \times \frac{W}{s}}$, where $K$ denotes the number of wheat disease categories. The maps are generated by a
light-weight generator $\mathcal{A}(\cdot)$ as follows:
\begin{equation}
    T = \sigma(\mathcal{A}(\tilde{F}_{D})),
\end{equation}
where $\sigma(\cdot)$ denotes the softmax activation function, $\mathcal{A}(\cdot)$ is a convolution with kernel size 1. 

Since each channel of $M$ explicitly corresponds to a specific disease category, we derive category-aware point prompt sets $P = \{P_1, P_2, \dots, P_K\}$. Specifically, we binarize these pattern maps using a threshold $t$ and extract the spatial coordinates of the activated regions, where $P_k$ denotes the dense positive prompts for the $k$-th disease category. Finally, the generated prompts in $\mathcal{P}$ are mapped from the DINOv2 feature grid back to the original image coordinate space, serving as category-specific dense positive prompts for SAM.

Therefore, the accurate generation of pattern maps plays a pivotal role in converting disease semantics into reliable point prompts. To ensure that these maps effectively capture diverse disease categories, we design an auxiliary loss to explicitly supervise them against the downsampled ground-truth masks $T_{gt}\in \mathbb{R}^{K \times \frac{H}{s} \times \frac{W}{s}}$. This constraint guarantees that the activated regions used for prompt extraction are semantically consistent with the actual disease boundaries:
\begin{equation}
    \mathcal{L}_{p} = \frac{1}{K}||T - T_{gt}||_2.
    \label{aux}
\end{equation}

Ultimately, SPG bridges the semantic-geometric gap by translating dense disease semantics into spatially precise point prompts. This transformation empowers SAM to effectively segment diseases characterized by extreme appearance variations. In contrast to labor-intensive manual prompting, PPG offers a fully automated, disease-sensitive strategy that unlocks SAM's potential for high-precision wheat disease segmentation.

\subsection{Geometric Feedback Segmentor}
\label{sec:gfs}

While dense category-aware prompts are essential for comprehensively covering complex wheat diseases, unconditionally decoding all candidates is computationally prohibitive and prone to accumulating noisy masks from imprecise locations. To mitigate this, we propose the Geometric Feedback Segmentor (GFS), which distills dense candidates into a compact yet highly accurate set of geometric guides. Specifically, GFS iteratively decodes category-specific prompts and progressively prunes low-quality prompts by cross-referencing SAM’s mask quality with DINOv2’s category-specific semantic cues.

Consistent with the DINOv2 branch, the input image $I$ is processed by a frozen SAM image encoder augmented with the proposed Disease Perception Adapter, denoted by $\mathcal{S}_{\mathcal{A}}(\cdot)$. The resulting disease-aware SAM feature is defined as
\begin{equation}
F_S = \mathcal{S}_{\mathcal{A}}(I),
\end{equation}
where $F_S$ preserves the geometric prior of SAM while enhancing its sensitivity to wheat disease structures.

Given the category-aware prompt sets $P$ produced by SPG, GFS performs category-wise iterative decoding to obtain semantic segmentation masks. For each category $k$, we initialize the remaining prompt pool as $R_k^0 = P_k$ and maintain a valid mask set $M_k = \emptyset$. At the $t$-th iteration, a point prompt batch $B^t \subseteq R_k^t$ is popped from the remaining prompt pool via uniform random sampling and fed into the SAM decoder together with $F_S$:
\begin{equation}
(\hat{M}_k, S_k) = \mathcal{S}_{\mathcal{D}}(F_S, B^t),
\end{equation}
where $\mathcal{S}_{\mathcal{D}}(\cdot)$ denotes the SAM decoder, $\hat{\mathcal{M}}_k \in \mathbb{R}^{|B^t|\times H \times W}$ is the set of candidate masks, and $\mathcal{S}_k \in \mathbb{R}^{|B^t|}$ denotes the corresponding mask quality scores predicted by SAM. This sampling strategy avoids redundant exhaustive decoding while maintaining sufficient coverage of the remaining lesion regions. 

However, masks decoded from sampled prompts may still be unreliable due to imprecise point locations or background interference. To obtain more reliable geometric feedback, we refine the mask selection process through a dual-validation strategy. Specifically, we jointly evaluate the geometric mask quality predicted by SAM and the semantics of the mask-guided DINOv2 features with the corresponding disease category.

Concretely, let $M_r \in \hat{M}_k$ denote the $r$-th candidate mask predicted by SAM, with its corresponding geometric quality score denoted by $s_{r}^{\mathrm{iou}} \in S_k$. To measure its semantic consistency with the target category $k$, we project the candidate mask onto the DINOv2 feature $\tilde{F}_D$ and compute a semantic score using a lightweight scoring network $\mathcal{F}_\mathcal{S}(\cdot)$:
\begin{equation}
  s_{r}^{\mathrm{sem}} = [\sigma( \mathcal{F}_\mathcal{S}(\tilde{F}_D \odot M_r) )]_k,  
\end{equation}
where $\sigma(\cdot)$ denotes the sigmoid function, and $[\cdot]_k$ extracts the probability score corresponding to category $k$. %By evaluating all masks in $\hat{\mathcal{M}}_k^t$, we obtain a corresponding set of semantic scores $\mathcal{S}_k^{\mathrm{sem}} = \{s_{1}^{\mathrm{sem}}, s_{2}^{\mathrm{sem}}, \dots, s_{|\hat{\mathcal{M}}_k^t|}^{\mathrm{sem}}\}$. Finally, the overall confidence score $s_r^{\mathrm{final}}$ for each candidate mask is computed by fusing its geometric quality and semantic consistency: $ \mathcal{S}^{final} = \{s_r^{\mathrm{iou}} \cdot s_r^{\mathrm{sem}}\}_{r\in[1, \cdots, |\hat{\mathcal{M}}_k^t|]}.$
Finally, by fusing the geometric quality and semantic consistency for all candidate masks in $\hat{\mathcal{M}}_k^t$, we obtain the overall confidence score set:
\begin{equation}
S_k^{\mathrm{final}} = \{ s_r^{\mathrm{iou}} \cdot s_r^{\mathrm{sem}} \}_{r=1}^{|\hat{M}_k|}.
\end{equation}
For each sampled prompt batch, we retain only the most reliable candidate mask:
\begin{equation}
r^\ast = \arg\max_r \mathcal{S}_k^{\mathrm{final}}(r), \qquad
(\hat{m}^k, s^k) = (\hat{m}_{r^\ast}^k, s_{r^\ast}^k).
\end{equation}
Finally, only masks whose confidence scores exceed a threshold $\tau$ are kept as valid predictions:
\begin{equation}
\tilde{M}_k = \left\{(\hat{m}^k, s^k)\mid s^k > \tau\right\}.
\end{equation}
The retained masks provide explicit geometric feedback for prompt pruning. Specifically, prompts that fall inside the currently accepted masks are regarded as redundant and removed from the remaining prompt pool:
\begin{equation}
R_k^{t+1}=R_k^t\setminus\left\{p \in R_k^t\mid\exists (\hat{m}, s) \in \tilde{M}_k,\; p \in \hat{m}\right\}.
\end{equation}
The valid mask-score pairs are then accumulated into the category-specific mask set:
\begin{equation}
M_k \leftarrow M_k \cup \tilde{M}_k.
\end{equation}
This process is repeated until $\mathcal{R}_k^t$ becomes empty. By progressively removing prompts already explained by confident masks, GFS focuses computation on unresolved disease regions and avoids redundant SAM decoding.

After iterative decoding, the valid masks of each category are aggregated to obtain the final semantic prediction. For the $k$-th category, the category-specific response map is computed in a pixel-wise manner as
\begin{equation}
\bar{M}_k(u,v)=\max_{(\hat{m},\, s)\in \mathcal{M}_k} \; s \cdot \hat{m}(u,v),
\end{equation}
where $\hat{m}(u,v)$ denotes the mask response at spatial location $(u,v)$, and the maximization selects the highest-confidence mask covering that location among all valid masks of category $k$. 

%In this way, GFS forms the geometric decoding stage of SGPer. Category-aware prompts from DINOv2 guide SAM toward disease-relevant regions, while the selected high-confidence masks provide geometric feedback to progressively refine the final lesion segmentation.

\subsection{Optimization Objective}
To supervise the $K$ category-specific response maps, we employ a joint segmentation loss comprising the Cross-Entropy and Dice losses:
\begin{equation}
    \mathcal{L}_{seg} = \mathcal{L}_{ce} + \mathcal{L}_{dice}.
\end{equation}
The overall training objective is defined as:
\begin{equation}
\mathcal{L} = \mathcal{L}_{seg} + \alpha \mathcal{L}_{p},
\label{loss}
\end{equation}
where $\alpha$ is a hyperparameter designed to weight the auxiliary loss term against the primary segmentation objective.

\section{Experiments}
\subsection{Experimental Setup}
\noindent\textbf{Datasets.} The Eschikon Foliar Disease v2 (EFDv2) dataset consists of 400 manually selected image crops of unstructured canopy, with a resolution of $1024 \times 1024$ pixels, divided into 320 images for training and 80 images for testing.
The Organ Segmentation Dataset (OSD) comprises 184 images of wheat canopies, with a resolution of $1024 \times 1024$ pixels, which is split into 147 images for training and 37 images for testing following an $80/20\%$ split.
The Global Wheat Full Semantic Organ Segmentation (GWFSS) \cite{wang2025global} dataset comprises a total of 308 labeled images. These images are divided into 99 images for training, 99 images for validation, and 110 images for testing.

\noindent\textbf{Implementation Details.} We use SAM (ViT-L)~\cite{kirillov2023segment} and DINOv2 (ViT-L)~\cite{oquab2024dinov2} as the default backbones in all experiments. To avoid overfitting in the limited-supervision setting, the pretrained parameters of both backbones are frozen, and only the proposed lightweight modules are updated during training. Supervision is provided by the ground-truth segmentation masks with standard segmentation losses. We optimize parameters using Adam~\cite{DBLP:journals/corr/KingmaB14} with a learning rate of $10^{-5}$, weight decay of $10^{-4}$, $\beta_1=0.9$, and $\beta_2=0.99$. All experiments are conducted on a single NVIDIA A100 GPU with a batch size of 1 for 2000 training iterations.

\noindent\textbf{Evaluation protocols.} We employ the Mean Intersection over Union ($IoU$) and the Mean $F_1$-score ($F_1$) to evaluate segmentation performance. Specifically, $IoU$ measures the average overlap between the predicted and ground truth masks across all categories. $F_1$ assesses the harmonic mean of precision and recall, providing a robust metric for both symptom detection and region classification.

\subsection{Ablation Experiments} 
Table~\ref{tab:ab} reports the ablation study of the proposed modules on the EFDV2 dataset. Starting from the baseline without Disease Perception Adapter (DPA) and Geometric Feedback Segmentor (GFS), the model achieves 47.6\% mIoU and 56.7\% $F_1$. After inserting DPA only into the DINOv2 branch, the performance rises markedly to 58.2\% mIoU and 64.9\% $F_1$, indicating that enhancing disease-aware semantic perception is effective for locating wheat disease regions under cross-stage appearance variation. Further introducing DPA into the SAM branch boosts the performance to 67.5\% mIoU and 75.1\% $F_1$, demonstrating that jointly adapting both DINOv2 and SAM is crucial for associating disease semantics with geometric priors. In addition, adding GFS further improves the performance to 72.5\% mIoU and 82.8\% $F_1$, achieving the best results among all settings. This demonstrates that iteratively decoding category-specific prompts and dynamically pruning redundant ones based on mask quality can effectively reduce unnecessary decoding while preserving reliable disease cues. As a result, a limited number of prompts can more effectively activate the geometric priors of SAM, leading to more accurate disease boundary localization and stronger robustness to appearance variations across growth stages. %Overall, the steady improvement across all configurations confirms that DPA and GFS are complementary: DPA strengthens disease-aware semantic and geometric representations, while GFS further translates these representations into accurate and efficient lesion segmentation.

\begin{table}[t]
    \centering
    \caption{Ablation study of the proposed auxiliary branches, \textit{i.e.}, Disease Perception Adapter (DPA) and Geometric Feedback Segmentor (GFS) on EFDV2 dataset.}
    \setlength{\tabcolsep}{8pt}
    \begin{tabular}{ccc|cc}
     \toprule
      DINO+DPA  & SAM+DPA & GFS & $IoU$&$F_1$ \\
      %%\addlinespace
      \toprule
         \xmark&\xmark&\xmark&47.6\%&56.7\%\\ 
         %%\addlinespace
          \cmark&\xmark&\xmark& 58.2\%&64.9\%\\
         % %\addlinespace
          \cmark&\cmark&\xmark& 67.5\%&75.1\%\\
          %%\addlinespace
           \cmark&\cmark&\cmark& 72.5\%&82.8\%\\
           \toprule
    \end{tabular}
    
    \label{tab:ab}
\end{table}

\begin{table*}[t]
  \centering
  \caption{Comparison of segmentation performance between the competitive methods and our proposed method.}
  \label{tab:results_comparison}
   \setlength{\tabcolsep}{9pt}
  \begin{tabular}{lcccccccccc}
    \toprule
    \multirow{2}{*}{Method} & \multicolumn{2}{c}{Background}& \multicolumn{2}{c}{Necrosis} & \multicolumn{2}{c}{Insect Damage} & \multicolumn{2}{c}{Powdery Mildew} & \multicolumn{2}{c}{Average} \\
    \cmidrule(r){2-3} \cmidrule(lr){4-5} \cmidrule(lr){6-7} \cmidrule(lr){8-9} \cmidrule(l){10-11}
     & $IoU$ & $F_1$ & $IoU$ & $F_1$ & $IoU$ & $F_1$ & $IoU$ & $F_1$ & $IoU$ & $F_1$ \\
    \midrule
    DDRNet \cite{DBLP:journals/tits/PanHSJ23} & 96.7\% & 98.4\% & 71.0\% & 84.6\%&56.2\%&64.2\%&9.4\%&10.0\%& 58.4\% &64.2\%\\
    %\addlinespace
    Mask2Former \cite{DBLP:conf/cvpr/ChengMSKG22} & 97.0\%& 98.7\% & 71.5\%&  82.5\%&   66.5\%& 76.2 & 14.6\%& 20.8\%&62.4\%&69.5\%\\
        %\addlinespace
    MaskFormer \cite{DBLP:conf/nips/ChengSK21} &96.9\%&98.9\%&70.5\%&79.4\%&66.0\%&75.0\%&17.6\%&18.5\% &62.7\%&67.9\% \\
    %\addlinespace
    SAN \cite{DBLP:journals/pami/XuZWHB23}& 96.5\%&98.2\%&70.5\%&84.4\%&59.9\%&70.0\%&16.3\%&17.3\%&60.8\%&67.5\%\\
    %\addlinespace
    SegNeXt \cite{DBLP:conf/nips/GuoLHLC022} &96.8\%&98.1\%&71.6\%&88.3\%&63.2\%&73.7\%&18.6\%&20.0\%&62.5\%&70.0\%\\
    %\addlinespace
    Segformer \cite{DBLP:conf/nips/XieWYAAL21} & 97.0\%&98.3\%&72.0\%&88.0\%&61.5\%&69.9\%&20.3\%&20.3\%&62.7\%&69.1\%\\
    GCNet \cite{DBLP:conf/cvpr/YangW0W25} &97.0\%&98.6\%&71.9\%&88.4\%&62.4\%&70.1\%&19.7\%&20.2\%&62.8\%&69.3\%\\
%\addlinespace
 \midrule
    \textbf{SGPer} & \textbf{97.1\%} & \textbf{98.5\%} & \textbf{73.7\%} & \textbf{84.8\%} & \textbf{71.7\%} & \textbf{83.5\%} & \textbf{47.4\%} & \textbf{64.3\%} & \textbf{72.5\%} & \textbf{82.8\%} \\
    \bottomrule
  \end{tabular}
\end{table*}

\begin{table*}[t]
  \centering
  \caption{Comparison of wheat organ segmentation performance on the OSD dataset.}
  \label{tab:osd_results}
   \setlength{\tabcolsep}{13pt}
  \begin{tabular}{lcccccccc}
    \toprule
    \multirow{2}{*}{Method}  & \multicolumn{2}{c}{Background} & \multicolumn{2}{c}{Ears} & \multicolumn{2}{c}{Stems} & \multicolumn{2}{c}{Average} \\
    \cmidrule(r){2-3} \cmidrule(lr){4-5} \cmidrule(l){6-7} \cmidrule(l){8-9}
     & $IoU$ & $F_1$ & $IoU$ & $F_1$ & $IoU$ & $F_1$ & $IoU$ & $F_1$ \\
    \midrule
    DDRNet \cite{DBLP:journals/tits/PanHSJ23} & 89.0\% &95.1\% &78.0\% &86.4\% &67.0\% &77.6\% &78.0\% &86.4\% \\
    %\addlinespace
    Mask2Former \cite{DBLP:conf/cvpr/ChengMSKG22} & 88.8\% &94.3\% &81.5\% &88.3\% &66.8\% &80.0\% &79.0\% &87.4\% \\
    %\addlinespace
    MaskFormer \cite{DBLP:conf/nips/ChengSK21} & 86.9\% &95.5\% &26.3\% &26.3\% &69.8\% &81.6\% &61.0\% &67.8\% \\
    %\addlinespace
    SAN \cite{DBLP:journals/pami/XuZWHB23} & 84.7\% &92.2\% &75.8\% &83.1\% &55.8\% &70.9\% &72.1\% &82.1\% \\
    %\addlinespace
    SegNeXt \cite{DBLP:conf/nips/GuoLHLC022} & 90.0\% &95.5\% &83.4\% &90.6\% &69.1\% &79.4\% &80.8\% &88.5\% \\
    %\addlinespace
    Segformer \cite{DBLP:conf/nips/XieWYAAL21} &89.9\% &95.3\% &83.5\% &90.3\% &68.4\% &79.7\% &80.7\% &88.4\%  \\
    %\addlinespace
     GCNet \cite{DBLP:conf/cvpr/YangW0W25} & 90.0\% &95.4\% &84.1\% &90.9\% &69.4\% &80.7\% &81.2\% &89.0\% \\
     \midrule
    \textbf{SGPer} & \textbf{90.1\%}&\textbf{94.0\%}&\textbf{86.3\%} & \textbf{92.7\% }& \textbf{70.7\%} & \textbf{82.8\%} & \textbf{82.4\%} & \textbf{89.8\%} \\
        \bottomrule
  \end{tabular}
\end{table*}

\begin{table*}[t]
  \centering
  \caption{Performance of the proposed method on the GWFSS Dataset for wheat organ segmentation.}
  \label{tab:gwfss_results}
   \setlength{\tabcolsep}{9pt}
  \begin{tabular}{lcccccccccc}
    \toprule
    \multirow{2}{*}{Method} & \multicolumn{2}{c}{Background} & \multicolumn{2}{c}{Head} & \multicolumn{2}{c}{Stem} & \multicolumn{2}{c}{Leaf} & \multicolumn{2}{c}{Average} \\
    \cmidrule(r){2-3} \cmidrule(lr){4-5} \cmidrule(lr){6-7} \cmidrule(lr){8-9} \cmidrule(l){10-11}
     & $IoU$ & $F_1$ & $IoU$ & $F_1$ & $IoU$ & $F_1$ & $IoU$ & $F_1$ & $IoU$ & $F_1$ \\
     \midrule
     DDRNet \cite{DBLP:journals/tits/PanHSJ23} & 65.99\% & 81.5\% &  74.6\% & 80.5\% & 8.9\% &  9.7\% & 77.5\% & 89.1\% & 56.7\% & 65.2\% \\
     %\addlinespace
     Mask2Former \cite{DBLP:conf/cvpr/ChengMSKG22} & 66.9\% & 80.7\% &  81.5\% & 86.0\% & 22.8\% & 29.5\% & 79.0\% & 89.6\% &  62.5\% & 71.4\% \\
      %\addlinespace
     MaskFormer \cite{DBLP:conf/nips/ChengSK21} & 63.9\% &75.6\% &  79.4\% & 83.5\% & 22.6\% & 33.6\% & 77.0\% & 89.5\% & 60.7\% & 70.6\%  \\
     %\addlinespace
     SAN \cite{DBLP:journals/pami/XuZWHB23} & 57.3\% & 67.3\% & 76.2\% & 83.5\% & 15.6\% & 20.0\% & 75.1\% & 90.4\% &  56.0\% & 65.3\% \\
      %\addlinespace
      SegNeXt \cite{DBLP:conf/nips/GuoLHLC022} & 66.9\% & 80.0\% &  82.6\% & 88.5\% & 21.8\% &  28.4\% & 78.6\% & 89.4\% & 62.5\% & 71.6\% \\
      %\addlinespace
      Segformer \cite{DBLP:conf/nips/XieWYAAL21}  &  68.8\% &82.9\% &   80.7\% & 85.8\% & 17.6\% & 20.3\% &  79.2\% & 89.8\% & 61.6\% & 69.7\% \\
       %\addlinespace
       GCNet \cite{DBLP:conf/cvpr/YangW0W25} & 68.4\% &82.0\% &80.2\% &86.9\% &22.4\% &32.3\% &80.1\% &90.6\% &62.8\% &73.0\% \\
    \midrule
    \textbf{SGPer}  & \textbf{77.6\%} & \textbf{87.4\%} & \textbf{81.9\%} & \textbf{90.0\% }& \textbf{40.3\% }& \textbf{57.5\%} & \textbf{83.4\%} & \textbf{91.0\%} & \textbf{70.8\%} & \textbf{81.5\%} \\
    \bottomrule
  \end{tabular}
\end{table*}

\subsection{Comparison with State-of-the-art Methods}
\noindent\textbf{Limited-data wheat disease segmentation.}
To evaluate the effectiveness of the proposed framework, we conducted a comparative analysis against several state-of-the-art segmentation models, including Transformer-based architectures (Mask2Former \cite{DBLP:conf/cvpr/ChengMSKG22}, Segformer \cite{DBLP:conf/nips/XieWYAAL21}) and specialized perception models (eg., SAN \cite{DBLP:journals/pami/XuZWHB23}). As reported in Table \ref{tab:results_comparison}, SGPer consistently achieves superior performance across all categories, attaining a significant $IoU$ of 72.5\% and an $F_1$-score of 82.8\%, which represents a 9.8\% and 12.8\% improvement over the strongest baseline, respectively.While most competitive methods achieve satisfactory performance on classes with abundant training samples (e.g., Necrosis), they struggle significantly on Powdery Mildew due to limited training data, with mIoU scores remaining below 21\%.
In contrast, our framework achieves a breakthrough $IoU$ of 47.4\% for this challenging category, effectively doubling the performance of previous SOTA methods. This substantial gain validates our core hypothesis by treating wheat disease segmentation under limited data as a coupled task of disease-specific semantic perception and disease boundary localization, enabling SGPer to overcome the limitations of data scarcity and intra-class temporal variation.
% Furthermore, the iterative decoding and dynamic pruning of dense category-specific prompts ensure robust boundary delineation and efficient mask prediction, demonstrating the framework's adaptability and precision in complex field environments.

\noindent\textbf{Limited-data wheat organ segmentation.}
To further validate the generalizability of our framework, we evaluate SGPer on the OSD dataset for wheat organ segmentation, comparing it against established benchmarks such as DDRNet \cite{DBLP:journals/tits/PanHSJ23}, Mask2Former \cite{DBLP:conf/cvpr/ChengMSKG22}, and Segformer \cite{DBLP:conf/nips/XieWYAAL21}. As summarized in Tab. \ref{tab:osd_results}, SGPer achieves state-of-the-art (SoTA) performance with an Average $IoU$ of 82.4\% and an $F_1$-score of 89.8\%, consistently surpassing all competitive methods.Notably, our framework achieves superior precision in segmenting complex structural organs, with 86.3\% IoU on Ears and 70.7\% IoU on Stems, surpassing the strong SegNeXt baseline by 2.9\% and 1.6\%, respectively.
While traditional architectures such as MaskFormer exhibit substantial performance drops on specific categories (e.g., 26.3\% IoU on Ears), SGPer maintains high stability across diverse morphological structures. This robustness stems from integrating DINOv2’s high-level semantic features with SAM’s geometric priors, enabling precise localization of organ boundaries even under the dense occlusion.
 %These results demonstrate that the proposed adapter-based prompt decoding mechanism is not only effective for disease lesion detection but also highly adaptable to broader agricultural scene parsing tasks.

The robustness of our proposed SGPer is further underscored on the more demanding GWFSS dataset. As demonstrated in Table~\ref{tab:gwfss_results}, SGPer consistently outperforms state-of-the-art baselines, achieving a superior 70.8\% $IoU$ and 81.5\% $F_1$-score. Notably, on the difficult Stem category, SGPer reaches a 40.3\% $IoU$, nearly doubling the performance of SegNeXt (21.8\%). This substantial margin stems from treating wheat disease segmentation under limited data as a coupled task of object semantic perception and disease boundary localization. By establishing a complementary paradigm between foundation models, SGPer leverages DINOv2 to provide robust, category-aware semantic priors that are resilient to appearance variations across growth stages, while simultaneously guiding SAM to exploit its geometric priors for precise localization.

\begin{table}[t]
    \centering\caption{ Evaluation on EFDV2 for models with different adapters. }
     \setlength{\tabcolsep}{15pt}
    \begin{tabular}{l|cc}
    
     \toprule
         Method&$IoU$ & $F_1$  \\
         \midrule
         LoRA \cite{hu2022lora}& 68.5\% & 75.3\%\\
         LoRAND \cite{yin20231}& 69.3\% & 76.7\%\\
         ADAPTFORMER \cite{chen2022adaptformer} & 70.1\% & 79.5\%\\
          \midrule
         \textbf{Our DPA} & \textbf{72.5\%} & \textbf{82.8\%} \\
         \bottomrule
    \end{tabular}
    
    \label{tab:adapter}
\end{table}

\begin{table}[t]
    \centering
    \caption{Quantitative performance on EFDV2 under different threshold values ($\tau$) for converting DINOv2 features into point prompts.}
    
    \begin{tabular}{l|ccccccc}
    \toprule
      Threshold $\tau$  & 0.1 &0.2 &0.3 &0.4&0.5 &0.6 \\
       \midrule
        $IoU$ & 70.3&72.0&72.5&70.7&.68.2&67.6\\
        \bottomrule
    \end{tabular}
    
    \label{tab:theshold}
\end{table}

\begin{table}[t]
    \centering
    \caption{Exploration of the impact of the auxiliary loss (Eq.~\ref{aux}).}
    \setlength{\tabcolsep}{16pt}
    \begin{tabular}{l|cc}
         \toprule
         Method & $IoU$ & $F_1$  \\
         \midrule
        SGPer w/o $\mathcal{L}_p$ & 70.0\% & 79.1\% \\
        \midrule
         \textbf{SGPer w/o $\mathcal{L}_p$} & \textbf{72.5\%} & \textbf{82.8\%} \\
          \bottomrule
    \end{tabular}
    \label{tab:aux_loss}
\end{table}

\subsection{Discussion} 
\noindent\textbf{Effect on diverse adapters.} We compare Disease-aware Perception Adapter (DPA) with several SoTA adaptation methods. As shown in Table \ref{tab:adapter}, our DPA significantly outperforms all baselines, achieving a superior 72.5\% $IoU$ and 82.8\% $F_1$. This performance improvement stems from DPA’s specialized design, as it enhances semantic perception of diverse disease patterns when integrated into DINOv2, and enables SAM to better exploit geometric priors for accurate modeling of irregular disease shapes.
By synergizing these domain-specific cues, DPA effectively resolves the challenge of extreme intra-class appearance variations across growth stages, which remains a bottleneck for general-purpose adapters.

\begin{table}[t]
    \centering\caption{ Quantitative performance of the model on EFDV2 when trained without or with dense point filtering on geometric feedback segmentor. }
    \setlength{\tabcolsep}{10pt}
    \begin{tabular}{l|cc}
     \toprule
         Method & $IoU$ & $F_1$  \\
         \midrule
         SGPer w/o Point Filtering & 69.4\% & 77.9\%\\
         \midrule
         \textbf{SGPer w Point Filtering} & \textbf{72.5\%} & \textbf{82.8\%}\\
        \bottomrule
    \end{tabular}
    \label{tab:point_filter}
\end{table}

\begin{table}[t]
    \centering
    \caption{Quantitative performance on EFDV2 dataset when the model is trained without DINOv2 features for mask quality assessment.}
    \setlength{\tabcolsep}{10pt}
    \begin{tabular}{l|cc}
    \toprule
         Method & $IoU$ & $F_1$  \\
        \midrule
         SGPer w/o DINOv2 features & 68.3\% & 76.2\%\\
         \midrule
         \textbf{SGPer w DINOv2 features} & \textbf{72.5\%} & \textbf{82.8\%}\\
        \bottomrule
    \end{tabular}
    \label{tab:dino}
\end{table}

\begin{table}[t]
    \centering
    \caption{ Quantitative performance of the model on EFDV2 when trained with different weights $\alpha$ in the loss function defined in Eq.~\ref{loss}.}
    \begin{tabular}{l|ccccccc}
    \toprule
       Weight $\alpha$  & 0.1 &0.2 &0.4 &0.6&0.8 &1.0  \\
       \midrule
        $IoU$ & 69.3&70.1&71.3&72.5&72.0&71.6\\
        \bottomrule
    \end{tabular}
    
    \label{tab:weight}
\end{table}

\noindent\textbf{Importance of threshold values.} We analyze the sensitivity of threshold $\tau$, which serves as a gating mechanism to filter DINOv2-derived prompts. As shown in Table~\ref{tab:theshold}, performance peaks at $\tau$=0.3 (72.5\% $IoU$). Lower thresholds ($\tau$<0.3) introduce excessive noise from non-object regions, degrading segmentation accuracy. Conversely, higher thresholds ($\tau$>0.3) lead to a sharp decline (\textit{e.g.}, 67.6\% at $\tau$=0.6) by filtering out critical points present in disease regions. For morphologically complex diseases, such sparse prompting fails to sufficiently activate SAM’s geometric perception, hindering precise boundary localization. Thus, $\tau$=0.3 provides the optimal balance between noise suppression and disease localization.

\noindent\textbf{Contribution of the auxiliary loss $\mathcal{L}_p$.} We conducted an ablation study as shown in Table \ref{tab:aux_loss}. Integrating $\mathcal{L}_p$ during the transformation of DINOv2 semantic features into SAM dense point prompts yields a significant improvement, increasing the $IoU$ from 70.0\% to 72.5\%. This enhancement stems from the explicit supervision provided by $\mathcal{L}_p$, which further constrains the generated prompts to precisely fall within wheat disease regions. By enforcing this spatial alignment, the semantic-to-prompt transition more effectively activates SAM's geometric priors, ensuring robust disease segmentation. These results confirm that $\mathcal{L}_p$ is essential for bridging the gap between high-level semantic perception and disease boundary localization in challenging field environments.

\noindent\textbf{Evaluation of the dense point filtering in Sec.~\ref{sec:gfs}.} We compare the performance of SGPer with and without the dense point filtering mechanism in Tab. \ref{tab:point_filter}. The integration of point filtering leads to a substantial improvement, boosting $IoU$ from 69.4\% to 72.5\% and $F_1$ by 4.9\%. This gain is primarily due to the effective mitigation of redundancy in the dense prompts converted from DINOv2 semantic features. By selectively filtering prompts that specifically contribute to disease segmentation, the model successfully eliminates points falling on background regions or distracting artifacts. This process prevents negative interference during the geometric decoding stage, ensuring that only high-confidence, disease-relevant prompts activate SAM's priors for precise boundary localization.

\noindent\textbf{Effect of DINOv2 features on mask quality assessment.} The importance of DINOv2 features in mask quality assessment is further validated on the EFDV2 dataset. As shown in Table \ref{tab:dino}, incorporating DINOv2 features leads to a significant performance leap, improving $IoU$ from 68.3\% to 72.5\% and $F_1$ by 6.6\%. Unlike standard approaches that rely solely on SAM's internal features for quality scoring, our framework leverages DINOv2’s rich semantic embeddings to provide a higher-level context for mask evaluation. This semantic-aware assessment enables more accurate discrimination of mask reliability, effectively filtering out imprecise segmentation results. Consequently, this synergy between semantic and geometric features ensures a more robust and refined final output in wheat disease segmentation.

\noindent\textbf{Hyperparameter analysis.}
We further investigate the sensitivity of the hyperparameter $\alpha$ in the loss function (Eq. \ref{loss}) to achieve an optimal balance between different optimization objectives. As illustrated in Table \ref{tab:weight}, the model's performance on EFDV2 follows a typical bell-shaped curve as $\alpha$ increases from 0.1 to 1.0. Specifically, the $IoU$ consistently improves from 69.3\% at $\alpha$=0.1 and reaches its peak at 72.5\% when $\alpha$=0.6. This trend suggests that a moderate weight for the corresponding loss term effectively guides the model to synergize semantic and geometric features. However, further increasing $\alpha$ beyond 0.6 leads to a gradual performance decline (e.g., 71.6\% at $\alpha$=1.0), likely due to the over-dominance of loss components which may suppress the disease boundary optimization. Consequently, we set $\alpha$=0.6 as the default configuration for all subsequent experiments.

\begin{figure}[!t]
    \centering
    \includegraphics[width=1\linewidth]{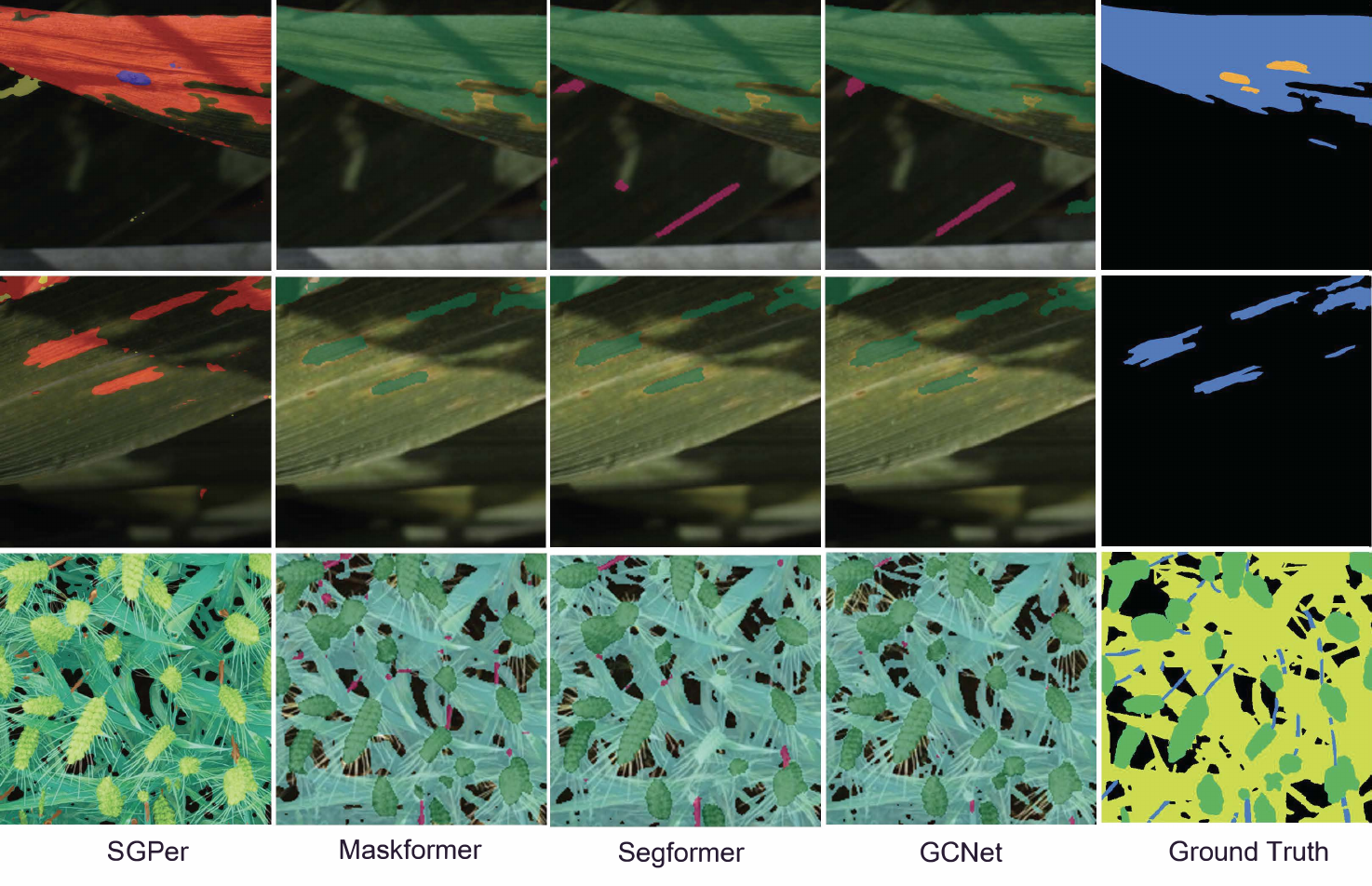}
    \Description{Qualitative comparison on the EFDV2 and GWFSS datasets.}
    \caption{Qualitative comparison for segmentation results. The first two rows present segmentation results on EFDV2, while the third row shows segmentation results on GWFSS.}
    \label{fig:qualitative_comparison}
\end{figure}
\begin{figure}[t]
    \centering
    \includegraphics[width=1\linewidth]{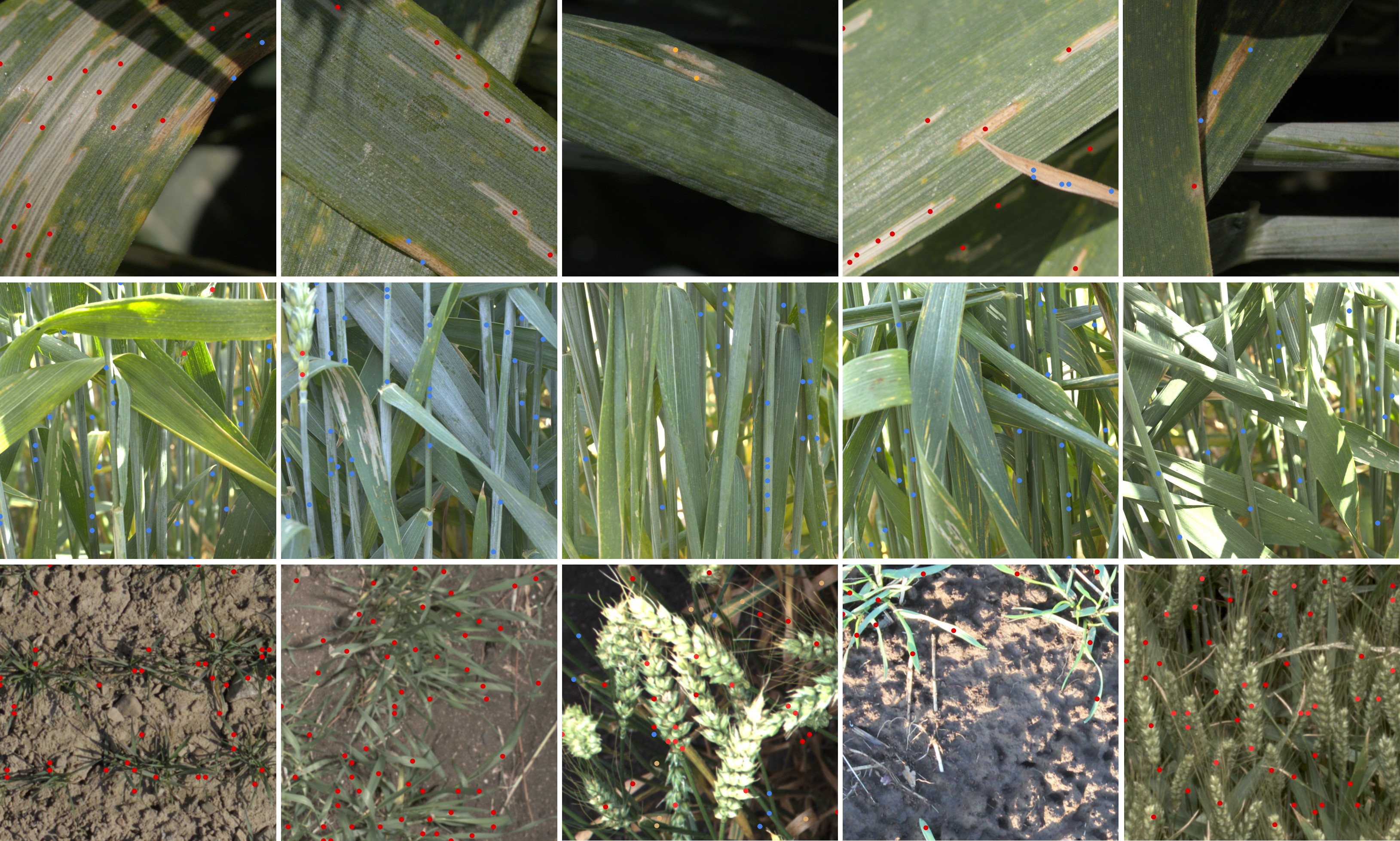}
    \caption{Visualization of point prompts across different samples. The first, second, and third rows correspond to EFDV2, OSD, and GWFSS, respectively. Best viewed with zoom.
}
    \label{fig:point}
\end{figure}

\noindent\textbf{Qualitative analysis.}
The qualitative results in Fig.~\ref{fig:qualitative_comparison} demonstrate the superiority of SGPer across the EFDV2 and GWFSS datasets. While baseline methods like GCNet and MaskFormer exhibit imprecise boundary localization due to high morphological variability, SGPer achieves precise localization of diverse diseases. This success is primarily attributed to the synergistic integration of DINOv2 and SAM: by transforming DINOv2’s high-level semantic representations into point-based prompts, SGPer effectively activates SAM’s disease boundary localization. Furthermore, for data-scarce categories such as stems in GWFSS, where prior methods struggle to learn discriminative features, SGPer leverages the strong generalization of both foundation models to maintain high accuracy. This collaboration enables the model to resolve both intra-class appearance variations and complex structural occlusions in challenging field environments.

\noindent\textbf{Visualization of point prompts.} 
As illustrated in Fig.~\ref{fig:point}, we visualize the generated point prompts across the EFDV2, OSD, and GWFSS datasets. Our framework effectively transforms semantic representations from DINOv2 into point prompts for SAM.Notably, SGPer exhibits strong adaptivity to object complexity, generating denser prompts for instances with complex contours or irregular appearances to ensure accurate boundary localization while retaining only a sparse set of informative prompts for objects with simpler structures. This dynamic allocation strategy effectively activates SAM’s geometric priors while reducing redundant computations, achieving a favorable balance between segmentation accuracy and inference efficiency.

\section{Conclusion} 
%In this paper, we propose SGPer, a semantic–geometric collaborative framework for wheat disease segmentation under limited supervision. It integrates the strong semantic representation capability of DINOv2 with the geometry-aware mask decoding of SAM, enabling disease semantics to guide geometric segmentation while geometric feedback further refines semantic perception. Consequently, SGPer effectively adapts foundation model priors to challenges arising from limited data and significantly intra-class temporal variations. Extensive experiments across multiple benchmarks demonstrate that SGPer consistently outperforms state-of-the-art methods, providing a robust and generalizable solution for precision agriculture.

In this paper, we propose SGPer, a novel Semantic-Geometric Prior Synergization framework designed to tackle the severe challenges of wheat disease segmentation under limited data and significant temporal appearance variations. By framing this challenge as a coupled task of semantic perception and geometric localization, SGPer successfully harmonizes the complementary strengths of DINOv2 and SAM. It introduces lightweight, disease-sensitive adapters to align generic foundation model priors with wheat disease characteristics. Furthermore, by translating DINOv2's robust semantic features into dense point prompts and employing a cross-referencing pruning mechanism, SGPer guarantees exhaustive spatial coverage while effectively filtering redundant geometric feedback. Extensive evaluations confirm that SGPer establishes a new state-of-the-art on wheat disease and organ segmentation benchmarks, demonstrating exceptional robustness in data-constrained scenarios.

\bibliographystyle{ACM-Reference-Format}
\bibliography{sample-base}


\end{document}